%% file: paper_a.tex
\documentclass[11pt]{article}
\usepackage[final]{acl}
\usepackage{times}
\usepackage{latexsym}
\usepackage[T1]{fontenc}
\usepackage[utf8]{inputenc}
\usepackage{microtype}
\usepackage{booktabs}
\usepackage{graphicx}
\usepackage{xcolor}
\usepackage{url}
\usepackage{amsmath}
\usepackage{array}
\usepackage{float}
\newcommand{\code}[1]{\texttt{#1}}
\newcommand{\RM}{\textsc{m}}
\newcommand{\RC}{\textsc{c}}
\input{generated/numbers}
\input{generated/threads}

\title{Train After You Measure: a Pre-Registered Learning Curve \\ for Fine-Tuning Typed Decision Models on One User's Labels}

\author{Federico Lucca \\
  Independent researcher, Trento, Italy \\
  \texttt{federico.lucca@gmail.com} \\
  ORCID \href{https://orcid.org/0009-0000-1493-6473}{0009-0000-1493-6473}}

\begin{document}
\maketitle
\begin{abstract}
Fine-tuning a small typed decision model on one user's email labels did not clear the margin we pre-registered over a logistic regression trained on the same labels. In earlier work, about two hundred labels and a TF-IDF logistic regression had the highest point accuracy on this inbox, though not significantly above the best hosted model; we ask whether the same labels, given to a decision model, close the gap. A first registered study, on a generic 4B base with cross-validation and message type only, stayed below the baseline at every label budget (\MLoraUoneMaxDiff{} to \MLoraUoneMinDiff{} points). The main study was registered before any fine-tuning run: a fixed test set of \TestRows{} emails never used for selection, LoRA adapters on the public Tev1-0.8B and Tev1-4B decision models, nested subsets of 50 to \PoolRows{} labels, two seeds and five decision fields on a free T4 GPU. Four of its five hypotheses are refuted---two in the fine-tunes' favour---and one is confirmed for one model. On message type the 4B model misses the registered three-point margin at 400 labels (\DiffLargeFourHundred) and passes it with all \PoolRows{} (\DiffLargeAll); with all labels both models are above the baseline by \DiffSmallAll{} and \DiffLargeAll{} points, where one email is \OneEmail{} points and a 95\% interval is about \WilsonWidth{} points wide, and the 4B model is clearly ahead in type macro-F1 (\TypeFAllLarge{} vs.\ \TypeFAllBase). On the three action switches the baseline is ahead at every label budget for both models (\SwAllBase{} vs.\ \SwAllSmall{} and \SwAllLarge{} mean accuracy with all labels), at under a millisecond per question against \LatSmall{} and \LatLarge{}\,ms. The 0.8B type curve barely improves after \SatSmall{} labels and then falls. The baseline never predicts the two rarest classes (\PoolOffer{} offers and \PoolSpam{} spam in the training pool); the fine-tuned models find spam in some runs, not consistently across seeds. We list the explanations we did not test and say when a new attempt would be worth making. Scripts are public; weights and emails are not.
\end{abstract}

\section{Introduction}

In earlier work we measured, on one owner's inbox, how well a personal email agent answers five typed questions about each message---its \emph{type}, its \emph{urgency}, and three switches that drive the agent (\emph{needs-reply}, \emph{action-48h}, \emph{is-noise})---against human labels collected before any training~\citep{lucca2026measure}. About two hundred labels and a TF-IDF logistic regression had the highest point accuracy on every field, ahead of every zero-shot typed decision model~\citep{jev,nimble} and generative LLM we tried, but with no significant lead over the best hosted model on any of them. The natural next move is to give the same labels to a decision model: fine-tune it with LoRA~\citep{hu2022lora} on the owner's answers, and see whether training closes the gap that zero-shot use left open.

We made that move twice, both times under a registration committed before any fine-tuning result was seen~\citep{nosek2018prereg}. Registration~\RM{} (30 October 2026, commit \code{56b2970}) adapted the generic Qwen3.5-4B base on message type, with thread-grouped cross-validation. Registration~\RC{} (\PreregDate, commit \code{96a3b7b}; Amendment~1, commit \code{3d88d63}) adapted the public Tev1-0.8B and Tev1-4B decision models~\citep{tev1small,tev1large} on all five fields, with nested label budgets from 50 to \PoolRows{}, two seeds, and a fixed test of \TestRows{} emails that was never used to choose anything. This paper reports both, with \RC{} as the main result. Both registrations are public at \url{https://github.com/federicolucca/foyer-measure/tree/v2.5/prereg}, verbatim in Italian with privacy redactions only; the commit hashes refer to the author's private repository, where they were first committed.

The answer, on this inbox, is: not by the margin we registered, and not where it matters most. On message type the fine-tunes end level with or slightly above the baseline (\DiffSmallAll{} and \DiffLargeAll{} points with all labels), inside a \WilsonWidth-point interval, and the 4B model is clearly better on the rare classes the baseline never predicts (macro-F1 \TypeFAllLarge{} vs.\ \TypeFAllBase); but they miss the registered three-point margin at 400 labels, lose to the baseline on the three switches that drive the agent's actions at every budget, and cost hundreds to thousands of times more per question. Four of the five hypotheses of~\RC{} are refuted, two of them in the fine-tunes' favour. We report every outcome as registered and offer the explanations only as untested hypotheses.

\paragraph{Related work.} Typed decision models answer a typed question with a distribution over its options~\citep{jev}; Nimble published an open recipe with LoRA on answer tokens~\citep{nimble}, and LLM-as-Jev discusses when such fine-tuning helps~\citep{llmasjev}. Linear classifiers on bag-of-words features remain strong text-classification baselines~\citep{lin2023linear}, prompts and fine-tuning trade off with the number of labelled points~\citep{lescao2021many}, and fine-tuning on few examples is unstable across seeds~\citep{dodge2020finetuning,mosbach2021stability}; learning curves are the standard tool to estimate how many labels a classifier needs~\citep{figueroa2012predicting}. Per-user email classification on a few hundred messages is an old problem~\citep{cohen96,klimt04,dredze08,grbovic14}. A longer discussion is in Appendix~\ref{app:related}.

\section{Setup}\label{sec:setup}

\paragraph{Data and split.} The labels are the owner's verified answers on the adversarial v3 corpus of~\citet{lucca2026measure}: \CorpusRows{} emails in \CorpusThreads{} threads, in Italian and English. Threads are taken from the corpus' thread field; \citet{lucca2026measure} grouped the same \CorpusRows{} emails into \ThreadsPaperOne{} threads with a normalised sender-plus-subject key (recomputed on the corpus, that key gives \GroupsSenderSubject{} groups). The stricter grouping used here is the one that defines the split, so no thread under either definition crosses train and test. Registration~\RC{} fixes one test set of \TestRows{} emails in \TestThreads{} threads (20\% of the threads of each type, seed 7) and leaves a pool of \PoolRows{} emails (\PoolThreads{} threads). From the pool, nested subsets of 50, 100, 200, 300, 400 and \PoolRows{} labels are drawn by whole threads in a type-stratified order, for seeds 7 and 11; a smaller subset is contained in every larger one, and no thread crosses train and test. The test holds \TestOffer{} \emph{offer} and \TestSpam{} \emph{spam} emails; the pool \PoolOffer{} and \PoolSpam{}.

\paragraph{Systems.} The \textbf{baseline} is a TF-IDF of word 1--2-grams and character 3--5-grams with a class-balanced logistic regression~\citep{pedregosa2011sklearn}, one model per field, on sender, subject and cleaned body, with fixed hyper-parameters. The \textbf{fine-tunes} are LoRA adapters (rank \LoraRank, $\alpha=\LoraAlpha$, dropout 0.05, all linear layers, learning rate $2\cdot10^{-4}$, effective batch \LoraBatch, at most \LoraMaxLen{} tokens) on Tev1-0.8B-experimental and Tev1-4B-experimental, one adapter per budget and seed trained on the five questions together, for at most \LoraEpochs{} epochs with early stopping on a \LoraValFrac\% thread-level validation split carved out of the budget. Prompts use Tev1's native typed format~\citep{tev1repo}: a fixed system instruction, the state (sender, subject, body), the Italian question and option definitions of~\citet{lucca2026measure}, and the answer as an option letter; the loss is on the answer only. Prediction reads the distribution over the option letters, renormalised, which reproduces the constrained typed call; confidence is the probability of the chosen option. Budget~0 is the same model without an adapter (zero-shot). All runs used free Kaggle T4 GPUs in fp16; no weights were kept.

\paragraph{Metrics and hypotheses.} On the fixed test, per field: accuracy, macro-F1, F1 per class, expected calibration error with 10 bins~\citep{guo2017calibration} and median latency per question. With two seeds we report their mean and compare systems only on the same seeds; a \emph{point} is 0.01 of accuracy, and one test email is \OneEmail{} points. Registration~\RC{}, committed on \PreregDate{} before any LoRA run and before the baseline was scored on the test, states five hypotheses with thresholds and decision rules (Table~\ref{tab:hyp}); a script decides them from the result files. Amendment~1, committed before any LoRA result was seen, withdrew a planned fallback to generic Qwen3.5 bases in place of Tev1: no Qwen base enters the curve. The 4B model ran budgets 0, 100, 400 and all; untested budgets count as not evaluable.

\paragraph{Registration \RM{}, and why its numbers are not \RC{}'s.} The preliminary study (commit \code{56b2970}, 30 October) differs on every axis that matters: 5-fold cross-validation by thread on all \CorpusRows{} labels, a word-only TF-IDF baseline, LoRA on the generic \code{Qwen/Qwen3.5-4B}, type only, two fixed epochs, prompts truncated to 768 tokens, free-text options. Its baseline reaches \MBaseKgUoneAccAll{} on type with all labels in cross-validation; \RC{}'s reaches \TypeAllBase{} on the fixed test. The two are not the same number and must not be compared, and neither is comparable with the $0.77\pm0.02$ of~\citet{lucca2026measure} (cross-validation on 231 v2 emails, five seeds of the split, against a test of \TestRows{} adversarially sampled v3 emails).

\section{Results}\label{sec:results}

\input{generated/tab_main}

\begin{figure*}[t]
\centering
\includegraphics[width=\textwidth]{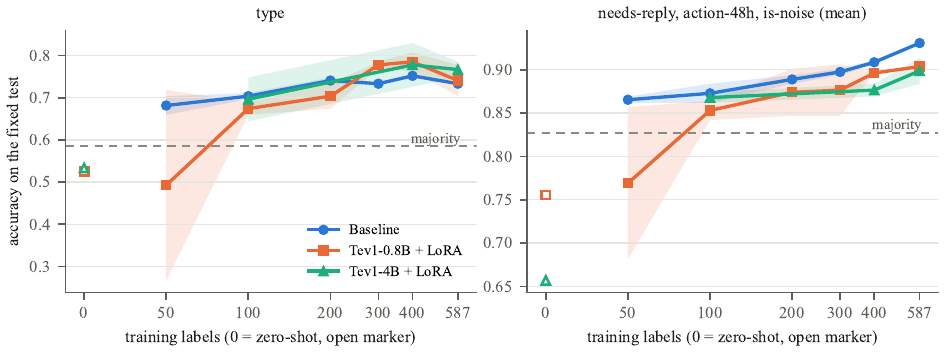}
\caption{Registration~\RC{}: accuracy on the fixed test by training labels (log scale), mean of seeds 7 and 11 with the min--max band across seeds; open markers are the same models zero-shot. Left: message type. Right: mean of the three switches. Dashed: majority class of the test. Generated from the result JSON files.}
\label{fig:curve}
\end{figure*}

\input{generated/tab_hyp}

\paragraph{Type.} The baseline is above the majority class from 50 labels on and reaches \TypeAllBase{} with all labels (Table~\ref{tab:main}, Figure~\ref{fig:curve}). Tev1-0.8B starts below the majority class at 50 labels, passes the baseline at 300 and 400 (\DiffSmallThreeHundred{} and \DiffSmallFourHundred{} points) and falls back to \DiffSmallAll{} with all labels. Tev1-4B is \DiffLargeHundred{} at 100, \DiffLargeFourHundred{} at 400 and \DiffLargeAll{} with all labels: H1$^c$ required at least +3 at every budget from 300 and is refuted at 400. From 100 labels on, every difference is at most six emails out of \TestRows{} on the mean of the seeds and inside the baseline's own 95\% Wilson interval~\citep{wilson1927probable} of [\WilsonBaseLo, \WilsonBaseHi]. The clearest gain is in macro-F1: with all labels the 4B model reaches \TypeFAllLarge{} against \TypeFAllBase{}, because it recognises the smaller classes better; this is a reported metric, not a registered hypothesis. Measured against each model's own zero-shot score on the seed where both exist, the labels close \GapClosedSmall\% (0.8B) and \GapClosedLarge\% (4B) of the distance to the baseline.

\paragraph{Switches and urgency.} On the mean of the three switches the baseline is ahead at every budget for both models; the fine-tunes' best is \SwBestDiffSmall{} (0.8B) and \SwBestDiffLarge{} (4B) points, and with all labels the gap is \SwAllBase{} vs.\ \SwAllSmall{} and \SwAllLarge{}. Per field (Appendix~\ref{app:tables}), neither fine-tune reaches the baseline on needs-reply; on urgency and action-48h the 0.8B model is ahead of it at some budgets between 100 and 400 labels, by fewer than three emails, and behind it with all labels; the 4B model is behind on both at every budget.

\paragraph{Saturation.} The 0.8B type curve gains \SlopesSmall{} points per 100 labels between consecutive budgets: the first slope below one point is at \SatSmall{} labels, so H2$^c$ holds for the 0.8B model, but the curve does not level off---it falls. ``Saturates'' here means that more labels stop helping. The 4B curve has too few budgets to evaluate H2$^c$.

\paragraph{Rare classes.} The baseline never predicts \emph{offer} or \emph{spam} (F1 \BaseRareMax{} at every budget and seed). The fine-tunes reach F1 $\geq0.3$ in \HFiveFtAgainst{} of \HFiveFtCases{} fine-tune cases (\HFiveBaseAgainst{} of \HFiveBaseCases{} baseline cases;\HFiveSpamAgainst{} on spam, \HFiveOfferAgainst{} on offer), which refutes H5$^c$; with \TestOffer{} offer and \TestSpam{} spam emails in the test, each such case is one or two emails, and only one (4B, spam, all labels) holds on both seeds (Appendix~\ref{app:tables}).

\paragraph{Calibration and cost.} Fine-tuning lowers the zero-shot ECE on type (\EceZeroSmall{} to \EceAllSmall{} for 0.8B, \EceZeroLarge{} to \EceAllLarge{} for 4B), but the 0.8B model stays at or above 0.10 at three of four budgets from 200 labels, refuting H4$^c$. A fine-tuned question takes a median \LatSmall{}\,ms (0.8B) or \LatLarge{}\,ms (4B) on a T4, against \LatBase{}\,ms for the baseline.

\paragraph{The preliminary study (registration \RM{}).} With the generic base, cross-validation and type only, the LoRA stayed below the baseline at every budget (in points: \MDiffs), with all labels \MLoraUoneAccAll{} vs.\ \MBaseKgUoneAccAll{} (Appendix~\ref{app:m}). H1$^m$ (+3 at 200 labels) was refuted at \MHoneDiff{} points; H2$^m$ was confirmed with the first slope below one point at \MHtwoSat{} labels, after which the curve also fell; H3$^m$ and H4$^m$ concerned a second user and are not evaluable; H5$^m$ was refuted, including by a case known at registration (Appendix~\ref{app:deviations}). The generic base did worse than the decision models, not better: it stayed below the baseline on type at every budget, where the decision models end level with it or slightly above.

\section{Why the margin was missed: untested explanations}\label{sec:why}

We did not test any of the following; they are ordered by how much we expect each to matter. (1) \emph{Too few examples of the rare classes}: the pool holds \PoolOffer{} offers and \PoolSpam{} spam, below the 15 per class that registration~\RM{} named as the threshold for learning them. (2) \emph{Large classes dominate the loss}: the baseline is class-balanced, the LoRA loss is not. (3) \emph{A small test}: with \TestRows{} emails the intervals are about \WilsonWidth{} points wide, wider than every difference on type. (4) \emph{Prompt format and length}: the baseline sees the whole text, the fine-tunes at most \LoraMaxLen{} tokens of a structured state. (5) \emph{Numerical precision}: fp16 on a T4. (6) \emph{The base}: a generic base (\RM) did worse than a decision model (\RC), so a better base is not obviously the answer. Testing them needs more labels per class (a second inbox), class weights in the LoRA loss, and a larger test set.

\section{What to do instead}\label{sec:instead}

On an inbox like this one, the baseline remains the system to serve: it is ahead on the three switches that trigger the agent's actions at every label budget, and it costs under a millisecond. On message type the fine-tunes are its equal within the test's resolution and better on the rare classes, so a fine-tune is a candidate for that field alone, if its latency is acceptable and the gain survives a larger test. Labels collected in production should go first to the baseline and to the contact graph of~\citet{lucca2026measure}, not to a fine-tune. A new fine-tuning attempt becomes worth its cost when every class has at least the registered 15 examples in the training pool and the test is large enough for the registered three-point margin to fall outside the baseline's interval.

\section*{Limitations}
One inbox and one user; labels written by the owner while reviewing model proposals, and therefore anchored on them; one LoRA hyper-parameter set for every budget, so the curve measures data, not tuning; two seeds, and one seed for the zero-shot points; a \TestRows{}-email test in which one email moves accuracy by \OneEmail{} points; and a 4B curve with four budgets.

\paragraph{Future work.} A second inbox, with the measurement kit of~\citet{lucca2026measure}; larger decision models as bases; and per-user versus shared fine-tuning. The protocol for a second user---per-user consents, per-user thread splits, checked before any label is read---is implemented and documented in the repository.

\paragraph{Reproducibility.} The scripts of the curve are in the public repository \url{https://github.com/federicolucca/foyer-measure} at tag v2.5, in \code{curve/}: \code{curve\_split.py} (split and subsets), \code{curve\_baseline.py}, \code{curve\_lora.py} (the Kaggle kernel) with \code{decision\_model.py}, \code{curve\_plot.py} and \code{curve\_tables.py} (every number, table and the figure of this paper), without data and without weights. The LoRA weights are not released, because adapters trained on email text can memorise it; the private Kaggle dataset was deleted after each run.

\section*{Ethics and Privacy}
No email text, sender, organisation or project name appears in this paper, its figure or its tables. The owner's consent to train on their own labels is stored as a file whose hash is recorded with every run, and the scripts check it before reading any label. Corpora uploaded to rented GPUs were deleted after each run, and no weights were kept there.

\paragraph{Acknowledgements.} We thank Haithem Afli (Munster Technological University) for his support.

\bibliography{refs}

\clearpage
\onecolumn
\raggedbottom
\appendix

\section{Deviations from the registrations}\label{app:deviations}

\paragraph{Base model (\RM).} Registration~\RM{} named Tev1 4B as the model to fine-tune. On 30 October the fine-tune was run on its generic base, \code{Qwen/Qwen3.5-4B}, on the belief that the Tev1 4B weights were not public; they were (checked on 5 November 2026), as \path{togethercomputer/Tev1-4B-experimental}~\citep{tev1large}. Registration~\RC{} uses the public Tev1 weights. The contrast is informative: the generic base stays below the baseline on type at every budget, while the decision models end level with it or slightly above on type and below it on the switches.

\paragraph{A known case against H5$^m$.} On the day of registration~\RM{}, before the commit, a trial baseline run reached F1 \MBaseUoneOfferFone{} on \emph{offer} with \MBaseUoneOfferN{} examples, above the 0.3 threshold although there are fewer than 15 offers. H5$^m$ was registered unchanged, so the case counts, and H5$^m$ could not be confirmed on any data including it. The trial run's type accuracy with all labels was \MBaseTrialAll{}; the baseline run paired with the LoRA on the same folds, used for every comparison, gives \MBaseKgUoneAccAll{} (different library versions).

\paragraph{Registration \RC.} Thread counts come from the corpus' thread field (\CorpusThreads{} threads); the option letter of Tev1's native format replaces the option key; the typed call is reproduced by constrained choice over the letters in Transformers, so latency is a forward pass on a T4, not a served call; the majority class of each training subset is reported besides that of the test. Amendment~1 is described in Section~\ref{sec:setup}.

\paragraph{Implementation (\RM).} The \RM{} curve was run in four Kaggle sessions; after the first, the prompt was encoded once per email and each option scored from a copy of its cache, logits were computed only at answer positions, and fast linear-attention kernels were used after a numerical check. Runs share data and split hashes and were merged without recomputing.

\section{Related work, longer}\label{app:related}
Jev~\citep{jev} defined the interface: a state, typed questions, and a distribution over the options of each. Open models followed, among them Nimble~\citep{nimble}, and Ollama exposed such models locally~\citep{ollama}. An evidence audit proposes an evaluation checklist~\citep{audit}, a benchmark compares Jev with general-purpose LLMs~\citep{bench}, and an ecosystem analysis maps its uses~\citep{wild}; on network-flow classification, classical classifiers trained on a few thousand records outperform Jev~\citep{traffic}. \citet{li26} show that most decision errors of System One models in agent loops fall on states read correctly, and confidence-gated escalation~\citep{li26judge,li26survey} motivates scoring calibration alongside accuracy. Email classification has long been personal: rule learning for filing~\citep{cohen96}, folder prediction on Enron~\citep{klimt04,bekkerman04}, reply prediction~\citep{dredze08}, and a handful of broad categories~\citep{grbovic14}; that work trained per-user classifiers on a few hundred messages, and the question it did not face is whether an adapted pre-trained decision model beats them at the same budget.

\section{Further tables}\label{app:tables}
\input{generated/tab_app_fields}
\input{generated/tab_app_calib}
\input{generated/tab_app_rare}

\clearpage
\section{Registration \RM{}}\label{app:m}
\input{generated/tab_app_m}

\end{document}

%% file: generated/numbers.tex
\newcommand{\CorpusRows}{722}
\newcommand{\CorpusThreads}{649}
\newcommand{\TestRows}{135}
\newcommand{\TestThreads}{129}
\newcommand{\PoolRows}{587}
\newcommand{\PoolThreads}{520}
\newcommand{\PoolOffer}{7}
\newcommand{\PoolSpam}{8}
\newcommand{\PreregDate}{5 November 2026}
\newcommand{\OneEmail}{0.74}
\newcommand{\TestOffer}{1}
\newcommand{\TestSpam}{2}
\newcommand{\LoraRank}{16}
\newcommand{\LoraAlpha}{32}
\newcommand{\LoraEpochs}{3}
\newcommand{\LoraMaxLen}{1024}
\newcommand{\LoraBatch}{16}
\newcommand{\LoraValFrac}{10}

\newcommand{\TypeAllBase}{0.733}
\newcommand{\TypeFAllBase}{0.406}
\newcommand{\SwAllBase}{0.931}

\newcommand{\SwAllSmall}{0.904}

\newcommand{\TypeFAllLarge}{0.581}
\newcommand{\SwAllLarge}{0.899}
\newcommand{\WilsonBaseLo}{0.653}
\newcommand{\WilsonBaseHi}{0.801}
\newcommand{\WilsonWidth}{15}

\newcommand{\DiffSmallThreeHundred}{+4.5}
\newcommand{\DiffSmallFourHundred}{+3.3}
\newcommand{\DiffSmallAll}{+0.7}
\newcommand{\SwBestDiffSmall}{$-$1.2}

\newcommand{\GapClosedSmall}{86}
\newcommand{\LatSmall}{306}
\newcommand{\DiffLargeHundred}{$-$0.7}
\newcommand{\DiffLargeFourHundred}{+2.6}
\newcommand{\DiffLargeAll}{+3.3}
\newcommand{\SwBestDiffLarge}{$-$0.5}

\newcommand{\GapClosedLarge}{107}
\newcommand{\LatLarge}{1681}
\newcommand{\LatBase}{0.7}
\newcommand{\SatSmall}{300}
\newcommand{\SlopesSmall}{36.3, 3.0, 7.4, 0.7, $-$2.4}
\newcommand{\BaseRareMax}{0.00}

\newcommand{\HFiveSpamAgainst}{6}
\newcommand{\HFiveOfferAgainst}{1}
\newcommand{\HFiveFtCases}{36}
\newcommand{\HFiveFtAgainst}{7}
\newcommand{\HFiveBaseCases}{24}
\newcommand{\HFiveBaseAgainst}{0}

\newcommand{\EceAllSmall}{0.126}
\newcommand{\EceZeroSmall}{0.263}
\newcommand{\EceAllLarge}{0.076}
\newcommand{\EceZeroLarge}{0.290}
\newcommand{\MLoraUoneAccAll}{0.734}
\newcommand{\MBaseKgUoneAccAll}{0.799}
\newcommand{\MLoraUoneMinDiff}{$-$7.3}
\newcommand{\MLoraUoneMaxDiff}{$-$1.8}
\newcommand{\MHoneDiff}{$-$1.8}
\newcommand{\MHtwoSat}{300}

\newcommand{\MBaseUoneOfferFone}{0.364}
\newcommand{\MBaseUoneOfferN}{8}

\newcommand{\MBaseTrialAll}{0.801}
\newcommand{\MDiffs}{50: $-$5.4, 100: $-$7.3, 200: $-$1.8, 300: $-$2.2, 400: $-$4.6, 500: $-$5.7, all: $-$6.5}

%% file: generated/threads.tex
\newcommand{\GroupsSenderSubject}{696}
\newcommand{\ThreadsPaperOne}{698}

%% file: generated/tab_main.tex
% GENERATED by curve_tables.py -- do not edit.
\begin{table*}[t]\centering\small
\caption{Registration C on the fixed test (135 emails). Accuracy and macro-F1 on \emph{type} and mean accuracy of the three switches, by training labels; mean of the seeds run at each budget (two everywhere except zero-shot, one). Budget 0 is the same model without LoRA. Majority class of the test: 0.585 on type, 0.827 on the switches. One email is 0.74 points; the 95\% Wilson interval of the baseline\textquotesingle s type accuracy with all labels is [0.653, 0.801]. Best value per row in bold. Generated by \texttt{curve\_tables.py} from the result JSON files.}
\label{tab:main}
\begin{tabular}{l ccc ccc ccc}\toprule
 & \multicolumn{3}{c}{type accuracy} & \multicolumn{3}{c}{type macro-F1} & \multicolumn{3}{c}{switches, mean accuracy} \\
\cmidrule(lr){2-4}\cmidrule(lr){5-7}\cmidrule(lr){8-10}
labels & base & 0.8B & 4B & base & 0.8B & 4B & base & 0.8B & 4B \\\midrule
0 (zero-shot) & -- & 0.526 & 0.533 & -- & 0.169 & 0.254 & -- & 0.756 & 0.657 \\
50 & \textbf{0.681} & 0.493 & -- & \textbf{0.338} & 0.161 & -- & \textbf{0.865} & 0.769 & -- \\
100 & \textbf{0.704} & 0.674 & 0.696 & 0.348 & 0.263 & \textbf{0.422} & \textbf{0.873} & 0.853 & 0.868 \\
200 & \textbf{0.741} & 0.704 & -- & \textbf{0.372} & 0.341 & -- & \textbf{0.889} & 0.874 & -- \\
300 & 0.733 & \textbf{0.778} & -- & 0.411 & \textbf{0.458} & -- & \textbf{0.898} & 0.877 & -- \\
400 & 0.752 & \textbf{0.785} & 0.778 & 0.431 & 0.442 & \textbf{0.536} & \textbf{0.909} & 0.896 & 0.877 \\
all (587) & 0.733 & 0.741 & \textbf{0.767} & 0.406 & 0.446 & \textbf{0.581} & \textbf{0.931} & 0.904 & 0.899 \\
\bottomrule\end{tabular}\end{table*}

%% file: generated/tab_hyp.tex
% GENERATED by curve_tables.py -- do not edit.
\begin{table}[t]\centering\footnotesize\setlength{\tabcolsep}{3pt}
\caption{Pre-registered hypotheses of registration C (committed 5 November 2026, before any LoRA run), decided by script from the result JSON files. Differences in accuracy points, fine-tune minus baseline, mean of the same seeds.}
\label{tab:hyp}
\begin{tabular}{@{}l >{\raggedright\arraybackslash}p{2.75cm} >{\raggedright\arraybackslash}p{2.75cm} >{\raggedright\arraybackslash}p{1.15cm}@{}}\toprule
 & registered prediction & observed & outcome \\\midrule
H1$^{c}$ & Tev1-4B beats the baseline on type by $\geq$3 points at every budget $\geq$300 & 400: +2.6; all: +3.3 & refuted \\
H2$^{c}$ & The fine-tuned type curve saturates at 300--400 labels (slope $<$1 point/100 labels) & 0.8B: first slope $<$1 at 300; 4B: curve incomplete & confirmed (0.8B only) \\
H3$^{c}$ & Tev1-0.8B does not beat the baseline on type at any budget & above at 300 (+4.5), 400 (+3.3), all (+0.7) & refuted \\
H4$^{c}$ & Type ECE of both fine-tunes $<$0.10 at every budget $\geq$200 & above at 0.8B 200 (0.109), 0.8B 400 (0.101), 0.8B all (0.126) & refuted \\
H5$^{c}$ & Offer and spam F1 $<$0.3 for every trained system, seed and budget & 7 of 36 fine-tune cases $\geq$0.3 (0 of 24 baseline cases; baseline max 0.00) & refuted \\
\bottomrule\end{tabular}\end{table}

%% file: generated/tab_app_fields.tex
% GENERATED by curve_tables.py -- do not edit.
\begin{table}[H]\centering\small\setlength{\tabcolsep}{4pt}
\caption{Registration C, the other four fields: accuracy (upper block) and macro-F1 (lower block) by training labels, mean of the seeds run; majority class of the test in the last row of each block.}
\label{tab:fields}
\begin{tabular}{l ccc ccc ccc ccc}\toprule
 & \multicolumn{3}{c}{urgency} & \multicolumn{3}{c}{needs-reply} & \multicolumn{3}{c}{action-48h} & \multicolumn{3}{c}{is-noise} \\
\cmidrule(lr){2-4}\cmidrule(lr){5-7}\cmidrule(lr){8-10}\cmidrule(lr){11-13}
labels & base & 0.8B & 4B & base & 0.8B & 4B & base & 0.8B & 4B & base & 0.8B & 4B \\
\midrule\multicolumn{13}{l}{\emph{accuracy}} \\
0 & -- & 0.141 & 0.393 & -- & 0.837 & 0.859 & -- & 0.815 & 0.748 & -- & 0.615 & 0.363 \\
50 & 0.811 & 0.793 & -- & 0.900 & 0.889 & -- & 0.885 & 0.881 & -- & 0.811 & 0.537 & -- \\
100 & 0.815 & 0.819 & 0.770 & 0.911 & 0.881 & 0.881 & 0.885 & 0.893 & 0.878 & 0.822 & 0.785 & 0.844 \\
200 & 0.830 & 0.844 & -- & 0.933 & 0.889 & -- & 0.893 & 0.904 & -- & 0.841 & 0.830 & -- \\
300 & 0.844 & 0.841 & -- & 0.922 & 0.889 & -- & 0.904 & 0.922 & -- & 0.867 & 0.819 & -- \\
400 & 0.848 & 0.863 & 0.830 & 0.937 & 0.881 & 0.878 & 0.907 & 0.922 & 0.889 & 0.881 & 0.885 & 0.863 \\
all & 0.867 & 0.852 & 0.837 & 0.963 & 0.915 & 0.904 & 0.926 & 0.915 & 0.915 & 0.904 & 0.881 & 0.878 \\
majority & \multicolumn{3}{c}{0.793} & \multicolumn{3}{c}{0.881} & \multicolumn{3}{c}{0.881} & \multicolumn{3}{c}{0.719} \\
\midrule\multicolumn{13}{l}{\emph{macro-F1}} \\
0 & -- & 0.107 & 0.316 & -- & 0.456 & 0.703 & -- & 0.486 & 0.521 & -- & 0.587 & 0.340 \\
50 & 0.353 & 0.295 & -- & 0.651 & 0.526 & -- & 0.525 & 0.469 & -- & 0.707 & 0.448 & -- \\
100 & 0.366 & 0.379 & 0.409 & 0.703 & 0.469 & 0.521 & 0.525 & 0.609 & 0.532 & 0.723 & 0.641 & 0.793 \\
200 & 0.423 & 0.470 & -- & 0.794 & 0.565 & -- & 0.599 & 0.620 & -- & 0.774 & 0.793 & -- \\
300 & 0.462 & 0.481 & -- & 0.763 & 0.579 & -- & 0.649 & 0.790 & -- & 0.818 & 0.691 & -- \\
400 & 0.481 & 0.496 & 0.445 & 0.816 & 0.469 & 0.515 & 0.654 & 0.744 & 0.526 & 0.838 & 0.857 & 0.838 \\
all & 0.517 & 0.482 & 0.474 & 0.903 & 0.779 & 0.648 & 0.753 & 0.692 & 0.794 & 0.871 & 0.856 & 0.855 \\
majority & \multicolumn{3}{c}{0.295} & \multicolumn{3}{c}{0.469} & \multicolumn{3}{c}{0.469} & \multicolumn{3}{c}{0.418} \\
\bottomrule\end{tabular}\end{table}

%% file: generated/tab_app_calib.tex
% GENERATED by curve_tables.py -- do not edit.
\begin{table}[H]\centering\small
\caption{Registration C, \emph{type}: ECE (10 bins), mean confidence on correct and wrong answers, median latency per question on a T4 (ms), best epoch per seed. Baseline median latency: 0.7\,ms.}
\label{tab:calib}
\begin{tabular}{llccccc}\toprule
model & labels & ECE & conf.\ right & conf.\ wrong & ms & epoch \\\midrule
0.8B & 0 & 0.263 & 0.854 & 0.716 & 269 & -- \\
0.8B & 50 & 0.224 & 0.701 & 0.679 & 310 & 3,1 \\
0.8B & 100 & 0.070 & 0.757 & 0.653 & 304 & 2,3 \\
0.8B & 200 & 0.109 & 0.834 & 0.716 & 305 & 2,3 \\
0.8B & 300 & 0.079 & 0.855 & 0.744 & 303 & 2,3 \\
0.8B & 400 & 0.101 & 0.886 & 0.756 & 304 & 3,2 \\
0.8B & all & 0.126 & 0.820 & 0.691 & 306 & 2,2 \\
4B & 0 & 0.290 & 0.856 & 0.786 & 1650 & -- \\
4B & 100 & 0.122 & 0.816 & 0.690 & 1692 & 1,2 \\
4B & 400 & 0.087 & 0.853 & 0.735 & 1694 & 2,2 \\
4B & all & 0.076 & 0.858 & 0.680 & 1681 & 2,3 \\
\bottomrule\end{tabular}\end{table}

%% file: generated/tab_app_rare.tex
% GENERATED by curve_tables.py -- do not edit.
\begin{table}[H]\centering\small
\caption{F1 on \emph{offer} / \emph{spam} on the fixed test (1 offer and 2 spam emails), per system and seed. -- = not run.}
\label{tab:rare}
\begin{tabular}{llcccccc}\toprule
system & seed & 50 & 100 & 200 & 300 & 400 & all \\\midrule
base & 7 & 0.00/0.00 & 0.00/0.00 & 0.00/0.00 & 0.00/0.00 & 0.00/0.00 & 0.00/0.00 \\
base & 11 & 0.00/0.00 & 0.00/0.00 & 0.00/0.00 & 0.00/0.00 & 0.00/0.00 & 0.00/0.00 \\
0.8B & 7 & 0.00/0.00 & 0.00/0.00 & 0.00/0.00 & 0.00/0.00 & 0.40/0.00 & 0.00/0.00 \\
0.8B & 11 & 0.00/0.00 & 0.00/0.00 & 0.00/0.00 & 0.00/0.67 & 0.00/0.00 & 0.00/0.33 \\
4B & 7 & -- & 0.00/0.00 & -- & -- & 0.00/0.00 & 0.00/0.67 \\
4B & 11 & -- & 0.00/0.67 & -- & -- & 0.00/1.00 & 0.00/0.67 \\
\bottomrule\end{tabular}\end{table}

%% file: generated/tab_app_m.tex
% GENERATED by curve_tables.py -- do not edit.
\begin{table}[H]\centering\small
\caption{Registration M (preliminary): LoRA on the generic Qwen3.5-4B base vs.\ a word TF-IDF baseline on \emph{type}, out-of-fold, same thread-grouped folds; $n$ = training labels per fold. Not comparable with Table~\ref{tab:main} (cross-validation, not the fixed test). Generated from the run\textquotesingle s \texttt{metrics.json}.}
\label{tab:m}
\begin{tabular}{llcccc}\toprule
$n$ & system & acc. & macro-F1 & ECE & ms/email \\\midrule
50 & LoRA & 0.634 & 0.298 & 0.232 & 8399.7 \\
50 & baseline & 0.688 & 0.329 & 0.351 & 0.3 \\
100 & LoRA & 0.665 & 0.319 & 0.151 & 2137.2 \\
100 & baseline & 0.738 & 0.379 & 0.406 & 0.3 \\
200 & LoRA & 0.745 & 0.481 & 0.135 & 2122.6 \\
200 & baseline & 0.763 & 0.413 & 0.423 & 0.3 \\
300 & LoRA & 0.758 & 0.455 & 0.111 & 2071.7 \\
300 & baseline & 0.780 & 0.443 & 0.429 & 0.4 \\
400 & LoRA & 0.744 & 0.439 & 0.093 & 2088.8 \\
400 & baseline & 0.789 & 0.511 & 0.411 & 0.3 \\
500 & LoRA & 0.733 & 0.434 & 0.162 & 2117.6 \\
500 & baseline & 0.789 & 0.543 & 0.388 & 0.4 \\
all & LoRA & 0.734 & 0.554 & 0.154 & 8393.3 \\
all & baseline & 0.799 & 0.565 & 0.378 & 0.3 \\
\bottomrule\end{tabular}\end{table}